\documentclass[sigconf,9pt]{acmart}
\usepackage{url}
\usepackage{amsmath,amsfonts}
\usepackage{algorithmic}
\usepackage{graphicx}
\usepackage{textcomp}
\usepackage{amsmath}
\usepackage{soul}
\usepackage{relsize}
\usepackage{adjustbox}
\usepackage{subfigure}
\usepackage{hhline}
\usepackage{array}
\usepackage{multirow}
\usepackage{tikz}
\usepackage{colortbl}
\usepackage{xspace}
\usepackage{marvosym}
\usepackage[titletoc,title]{appendix}
\newcommand{\ciao}[1]{{\setlength\fboxrule{0pt}\fbox{\tcbox[colframe=black,colback=white,shrink tight,boxrule=0.8pt,extrude by=1mm]{\small #1}}}}

\usepackage{xspace}

\newcommand\nth{\textsuperscript{th}\xspace} 
\newcommand\nst{\textsuperscript{st}\xspace} 

\def\BibTeX{{\rm B\kern-.05em{\sc i\kern-.025em b}\kern-.08em
    T\kern-.1667em\lower.7ex\hbox{E}\kern-.125emX}}

\newcommand{\myverb}{\fontsize{9}{48}\usefont{OT1}{lmtt}{b}{n}\noindent }

\newcommand{\ie}{{\em i.e., }}
\newcommand{\eg}{{\em e.g., }}
\usepackage{tcolorbox}

\setcopyright{none}

\begin{document}

\title[Impact of Concept Drifts on IoT Traffic Inference  in Residential ISP Networks]{Quantifying and Managing Impacts of Concept Drifts on\\ IoT Traffic Inference  in Residential ISP Networks}

\author[A. Pashamokhtari, N. Okui, M. Nakahara, A. Kubota, G. Batista and H. Habibi Gharakheili]{Arman Pashamokhtari$^\ddagger$ \hspace{1em} Norihiro Okui$^\star$ \hspace{1em} Masataka Nakahara$^\star$ \hspace{1em} Ayumu Kubota$^\star$ \hspace{1em}\\Gustavo Batista$^\ddagger$ \hspace{1em} Hassan Habibi Gharakheili$^\ddagger$}
\affiliation{
	\institution{$^\ddagger$UNSW Sydney, Australia \hspace{1em} $^\star$KDDI Research, Inc., Saitama, Japan}
	\country{}
}
\affiliation{
	\institution{Emails:\hspace{0.2em}\{a.pashamokhtari,\hspace{0.2em}g.batista,\hspace{0.2em}h.habibi\}@unsw.edu.au,\hspace{0.2em}\{no-okui,\hspace{0.2em}ms-nakahara,\hspace{0.2em}ay-kubota\}@kddi.com}
	\country{}
}

\begin{abstract}
Millions of vulnerable consumer IoT devices in home networks are the enabler for cyber crimes putting user privacy and Internet security at risk. Internet service providers (ISPs) are best poised to play key roles in mitigating risks by automatically inferring active IoT devices per household and notifying users of vulnerable ones. Developing a scalable inference method that can perform robustly across thousands of home networks is a non-trivial task. This paper focuses on the challenges of developing and applying data-driven inference models when labeled data of device behaviors is limited and the distribution of data changes (concept drift) across time and space domains.  
Our contributions are three-fold: (1) We collect and analyze network traffic of 24 types of consumer IoT devices from 12 real homes over six weeks to highlight the challenge of temporal and spatial concept drifts in network behavior of IoT devices; (2) We analyze the performance of two inference strategies, namely ``\textit{global inference}'' (a model trained on a combined set of all labeled data from training homes) and ``\textit{contextualized inference}'' (several models each trained on the labeled data from a training home) in the presence of concept drifts; and (3) To manage concept drifts, we develop a method that dynamically applies the ``closest'' model (from a set) to network traffic of unseen homes during the testing phase, yielding better performance in $20$\% of scenarios.


\end{abstract}
\keywords{traffic inference, concept drift, machine learning, IoT, IPFIX}

%
\maketitle

\section{Introduction}
IoT devices bring a number of vulnerabilities that make them attractive to cyber criminals. Weak (or default) passwords and open insecure service ports are considered the top two exploited vulnerabilities of IoT devices \cite{minim2020}. Vulnerable IoT devices are now becoming a risk challenge for ISPs \cite{bitdefender2020}. There exist commercial solutions like \cite{bitdefenderSolution, aviraSafeThings} that help ISPs tackle the challenge of vulnerable IoT devices. The first step in every risk assessment and security analysis is obtaining visibility \cite{TMC19arunan} by discovering IoT devices connected to the network and determining their type (\eg camera, TV, speaker) \cite{Palo2020} using  characteristics known a priori.

IoT devices often show identifiable patterns in their network behaviors, making them relatively distinguishable from each other (though behavioral overlaps are common too \cite{TMC19arunan}). However, these patterns may change \cite{detectChange2020,TNSM2021,TMA2021revistIoTclass} by the time and context of their use across various networks. The behavioral change is more pronounced when IoT traffic inference is the objective of an ISP tasked to serve and manage tens of thousands of home networks, each with a unique composition of assets and users, all distributed across sizable geography (city, state, or country). Several existing research works studied different methods that ISPs can leverage to detect IoT devices in residential networks \cite{LCN2021,Marchal19,Meidan2020,Miettinen2017,Thangavelu2019}. 
Prior works tend to train a global model with machine learning algorithms and fine-tuned it by traffic data (testing and training) collected from a testbed (representing a single context). Due to the limitation of data or evaluation scenarios, they did not encounter context variations and could not highlight and/or address their impacts.

Context-aware or contextualized modeling is an alternative to global modeling that consists of three stages: (1) a set of training contexts are identified; (2) models are trained on a per-context basis; and, (3) a given testing sample is predicted by the ``closest'' model selected from all available trained models \cite{Reis2018a, Reis2018b,Nasc2018}. Note that each contextualized model is trained by data collected from a single context (narrow and relatively tighter knowledge). In contrast, the global model captures data from multiple contexts (broad and relatively loose knowledge). Limiting a model to learn a single context may increase the chance of over-fitting, while exposing a model to a diversified data set may not necessarily result in better performance, especially when the data is noisy. Though contextualized modeling has been studied in other domains \cite{Reis2018a, Reis2018b,Nasc2018}, to the best of our knowledge, no relevant study is found in the area of IoT traffic inference.


This paper compares global and contextualized modeling for classifying IoT devices in home networks. Specifically, we aim to answer the following question: ``Given a labeled dataset (training) from $N$ homes 
and an unlabeled dataset (testing) from other $M$ homes, 
which of the global versus contextualized modeling does yield better performance in classifying devices during the testing phase?'' Our \textbf{first} contribution highlights the presence of concept drifts in IoT traffic behavior by analyzing more than 6M flow records (\S\ref{sec:difts}). For our \textbf{second} contribution, we develop global and contextualized models (aiming to manage concept drifts in the space domain) and compare their performance (\S\ref{sec:strategies}). For the \textbf{third} contribution, we demonstrate that a dynamic inference can be applied to a combination of global and contextualized models to address concept drifts in the time domain (\S\ref{sec:dynamic}).

\section{Concept Drifts in Network Behavior of IoT Devices}\label{sec:difts}
A prerequisite step for an ISP that aims at automatically detecting IoT devices in home networks is to collect labeled data on network traffic for an intended set of devices, generating one or more inference models. The ISP can encourage a \textit{limited} number of its subscribers to voluntarily (or by other incentives like discounting monthly bills) help with the labeling network traffic of their home. For example, software applications like ``IoT Inspector'' \cite{IoTInspector} can be installed on a computer (or home gateway) in each home and let users label the devices. Note that the data collection and labeling methods are beyond the scope of this paper. Nevertheless, the more homes the ISP can collect labeled data from, its inference method becomes richer. However, collecting labeled data is expensive in both time and cost for the ISP, so this can only be done for a limited number of home networks. 

Another approach for collecting the labeled data is that the ISP constructs a testbed in their lab, consisting of a handful of IoT devices (those vulnerable devices that they aim to detect), and collects labeled data from the testbed. The testbed approach is relatively easier, but the data quality may be synthetically higher than the data collected from real homes. Additionally, collecting traffic from multiple homes provides the ISP with multiple contexts of traffic data. In contrast, the testbed data only includes a single context that may be inadequate for thousands of new homes. In this paper, we use data collected from real home networks. 

Concept drift is a known challenge of inference models that occurs when the distribution of data changes in the testing set from what it was in the training set. This challenge can be perceived differently in the time domain (when the model was trained on data collected some time ago) versus the space domain (when the model was trained on data of different homes with slight variations in context).

In the scope of this paper (inferring IoT devices in home networks), we encounter both sources of concept drifts. The probability and intensity of concept drifts vary by how the ISP trains its inference models (composition of homes) and how often they get re-trained. Frequent re-training can make the models more tolerant to concept drifts in the time domain; however, the re-training process has its own practical challenges. Another point to consider is that the ISP may need to carefully select the training context (composition of homes) to cater to diversity, becoming relatively resilient to concept drifts in the space domain. 


\subsection{Data Collection}\label{sec:data}
For this paper, we collect network traffic from 12 real homes. We have a set of 24 IoT device types (makes and models), including two cameras, four power switches, two humidifiers, an air sensor, two speakers, two media streamers, three hubs, two lightbulbs, a weighing scale, a tablet {\color{black}(for controlling IoT devices through their respective application)}, a printer, a sleep sensor, a smart remote, and a vacuum cleaner. For each of these types, we procured 12 units, meaning a total of 288 IoT units.  Individual homes are given a unit of each device type. In other words, each of the 12 homes has its own set of 24 devices. We collected data from these homes for 47 days.

For an ISP serving thousands of households, the choice of traffic data and the measurement location are nontrivial tasks mainly due to operational and scalability challenges. Works in \cite{LCN2021, Saidi2020, Meidan2020} showed that these challenges could be managed by collecting data at the edge of the ISP network (outside homes). This paper also considers that traffic is measured at the same vantage point and that flow records are employed to manage computing costs. That said, it is important to note that to evaluate the efficacy of our inference methods, we collect raw packets (in PCAP format) from inside home networks to obtain true device labels. We next transform them into  post-NAT IPFIX records \cite{ipfixRFC5103} (emulating a real scenario) and use them for training and testing models.

\begin{table}[t!]
	\caption{Summary of our dataset.}
	\vspace{-3mm}
	\renewcommand{\arraystretch}{1.1}
	\begin{adjustbox}{width=0.3\textwidth}
		\begin{tabular}{|l|c|}
			\hline
			{\textbf{IoT device type}} & {\textbf{\# IPFIX records}}\\ \hline
			Google Nest & 1,980,998\\  \hline
			Google Chromecast & 916,243\\  \hline
			Amazon Echo & 632,296\\  \hline
			Amazon Fire TV remote & 526,872\\  \hline
			Atom camera & 523,321\\  \hline
			Amazon Fire 7 tablet & 368,749\\  \hline
			Switchbot humidifier & 246,439\\  \hline
			TP-Link camera & 215,201\\  \hline
			Qrio hub & 209,668\\  \hline
			Panasonic home unit & 107,225\\  \hline
			Switchbot hub & 97,608\\  \hline
			TP-Link plug& 79,001\\  \hline
			Switchbot plug& 71,269\\  \hline
			iRobot roomba & 61,598\\  \hline
			LinkJapan eSensor & 54,012\\  \hline
			Meross plug & 40,792\\  \hline
			Meross lightbulb & 36,373\\  \hline
			TP-Link lighbulb & 30,820\\  \hline
			Meross plug & 29,665\\  \hline
			Meross remote& 27,224\\  \hline
			Meross humidifier & 25,825\\  \hline
			Withings sleep sensor & 15,068\\  \hline
			Canon printer & 6,718\\  \hline
			Elecom scale & 2,641\\  \hline
			
		\end{tabular}
	\end{adjustbox}
	\label{tab:records}
\end{table}

Our entire dataset consists of a total of 6,305,626 IPFIX records shown in Table~\ref{tab:records} for each device. 
Each IPFIX record corresponds to a five-tuple flow which is distinguished with source/destination IP address, source/ destination port number, and IP protocol number. IPFIX records are bidirectional which means they include activity features corresponding to both directions of a flow \eg incoming packet count and outgoing packet count. Inspired by prior work, Table~\ref{tab:features} shows 11 activity features that we extract from each direction of IPFIX records. As IPFIX records are bi-directional, for each of these features, there is a reverse equivalent that captures activity of the other direction of the flow; hence, there is a total of 22 activity features. In addition to the activity features, we extract 6 binary features for indicating the protocol of flows into one of the following protocols: {\myverb{HTTP}}, {\myverb{TLS}}, {\myverb{DNS}}, {\myverb{NTP}}, {\myverb{TCP}} (other than {\myverb{HTTP}} and {\myverb{TLS}}), and {\myverb{UDP}} (other than {\myverb{DNS}} and {\myverb{NTP}}). Therefore, we extract a total of 28 features per IPFIX record. 

\begin{table}[t!]
	\caption{Activity features of IPFIX records.}
	\vspace{-2mm}
	\renewcommand{\arraystretch}{1.2}
	\begin{adjustbox}{width=0.475\textwidth}
		\begin{tabular}{|l|l|}
			\hline
			\textbf{Feature} & \textbf{Description}\\ \hline 
			\myverb{packetTotalCount}& \# packet\\ \hline 
			\myverb{octetTotalCount} & \# byte\\ \hline 
			\myverb{smallPacketCount} & \# packet with $<$ 60 bytes payload\\ \hline
			\myverb{largePacketCount} & \# packet with $\ge$ 220 bytes payload\\ \hline
			\myverb{nonEmptyPacketCount} & \# packet with payload\\ \hline
			\myverb{dataByteCount} & payload size in total\\ \hline 
			\myverb{averageInterarrivalTime} & packets inter-arrival time $\mu$\\ \hline 
			\myverb{firstNonEmptyPacketSize} & first non-empty payload size\\ \hline 
			\myverb{maxPacketSize} & maximum payload size \\ \hline 
			\myverb{standardDeviationPayloadLength} & payload size $\sigma$ \\ \hline
			\myverb{standardDeviationInterarrivalTime} & packets inter-arrival time $\sigma$\\ \hline
		\end{tabular}
	\end{adjustbox}
	\label{tab:features}
\end{table}

In this paper, we use multi-class Random Forest to develop our inference models as it has proven effective in network traffic inferencing \cite{zhao21}. Model inputs are IPFIX features, and outputs are a device type (from 24 classes) along with a classification score. We note that the inference is not bound to any specific algorithm, so one may choose to use other methods like neural networks for this purpose. As IPFIX records are coarse-grained, it is not unlikely for the model to be less confident in its prediction. To increase the prediction quality, we apply class-specific thresholds to the score given by the model, thereby accepting predictions accompanied by relatively high scores. We obtain class-specific thresholds during the training phase by taking the average score for correctly predicted training instances per class.

\subsection{Temporal Drifts in IoT Behaviors}\label{subsec:timedrift}
%

Let us begin with how the behavior of IoT devices changes over time. {\color{black} Temporal drifts often can be attributed to automatic software/firmware upgrades by the manufacturer of devices \cite{TNSM2021}}. We use the first 30 days of our collected data for training and the remaining 17 days for testing. We train a Random Forest classifier for each home using its corresponding training dataset, obtaining 12 inference models \ie one model per home ($m_i: i\in[1,12]$).

\begin{figure}[t!]
	\centering
	\includegraphics[width=0.995\linewidth]{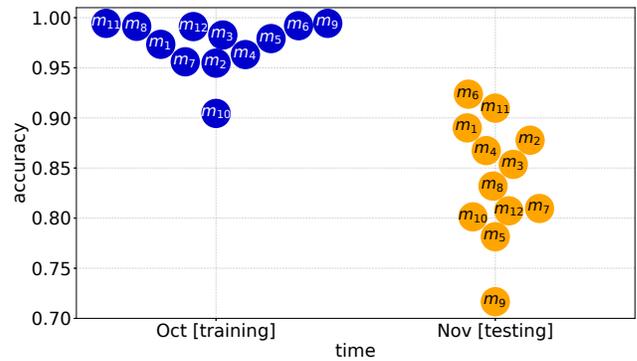}
	\caption{Temporal drifts in IoT network behaviors lead to performance decay across models (one per each home) from training to testing.}
	\label{fig:temporal drift}
\end{figure}

We apply these 12 models to their corresponding training and testing data to track their performance (average prediction accuracy across 24 classes). Fig.~\ref{fig:temporal drift} shows the accuracy (a value between 0 and 1) of models on average drops from training (shown in blue circles) to testing (orange circles).
All models realize a lower accuracy in the testing phase than in the training phase. Temporal drifts  deteriorated the performance of models by about $13$\% on average. The smallest and largest gaps are seen for $m_6$ and $m_9$, where their testing accuracy drops by more than $6$\% and $27$\%, respectively.


\subsection{Spatial Drifts in IoT Behaviors}

\begin{figure}[t!]
	\centering
	\includegraphics[width=0.995\linewidth]{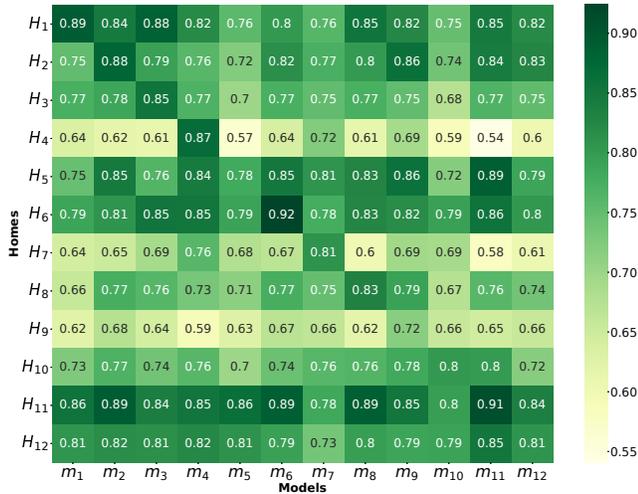}
	\vspace{-2mm}
	\caption{Spatial drifts in IoT network behaviors deteriorate model performance when tested against data of other contexts.}
	\vspace{-2mm}
	\label{fig:spatial drift}
\end{figure}

The second form of drift is when models learn and infer across the spatial domain, {\color{black}where users configure and operate their IoT devices differently \eg a smart camera configured with high-resolution settings will generate larger volumes of network traffic than a low-resolution one}.
To analyze this scenario, we apply the models ($m_i: i\in[1,12]$) trained in \S\ref{subsec:timedrift} to the testing set of all homes ($H_i: i\in[1,12]$). Fig.~\ref{fig:spatial drift} shows the results of this experiment. Each cell is the average prediction accuracy of models (listed across columns) when tested against data of homes (listed across rows). 

Overall, models perform better when they apply to their corresponding context. We observe that diagonal cells are often among the highest in each row, with some exceptions. For example, in $H_{12}$, its own model $m_{12}$ is beaten by $m_{11}$, giving an accuracy of $0.81$ versus $0.85$. Such an unexpected pattern is more pronounced in $H5$, where the performance of $m_5$ is lower than that of eight other models.   

It can be seen that certain homes like $H_{11}$ and $H_{12}$ receive relatively consistent predictions (fairly green cells) from all models. Conversely, in homes like $H_4$ and $H_7$ inconsistent performance across models is evident (yellow versus green cells) -- spatial drifts deteriorate the performance more. Considering $H_4$, for example, the accuracy of its own model ($m_4$) is $0.87$, whereas all other models at best give an accuracy of $0.72$ -- a non-negligible gap of $0.15$.  


\textbf{Summary:} Concept drifts are unavoidable, particularly at the scale of an ISP with only a limited amount of labeled data available from a diverse set of homes. In the following sections, we will develop strategies and methods to manage the impact of concept drifts on our IoT traffic inference.

\section{IoT Traffic Inference Strategies}\label{sec:strategies}

\begin{figure}[t!]
	\centering
	\includegraphics[width=0.995\linewidth]{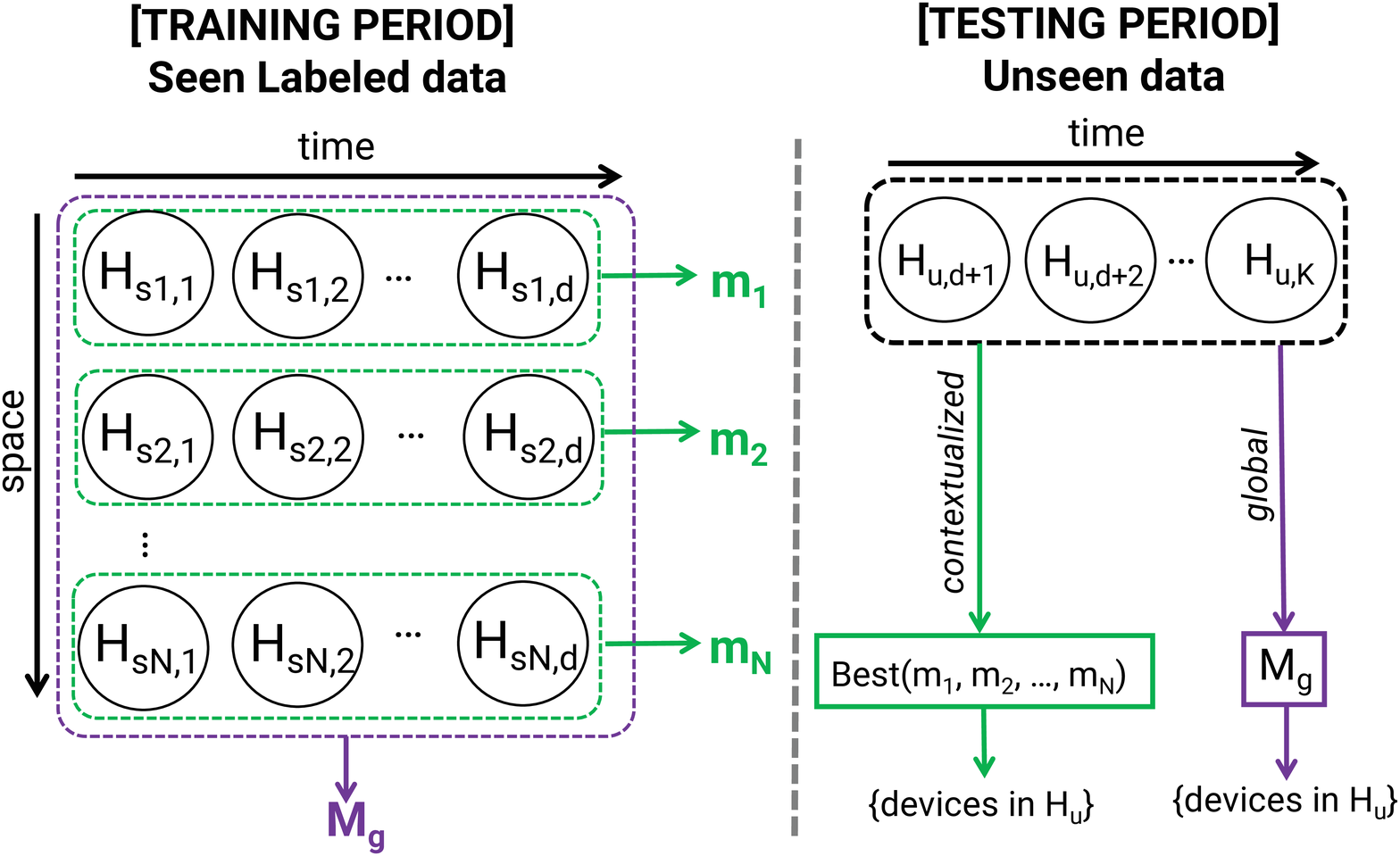}
	\vspace{-2mm}
	\caption{Global versus contextualized modeling for detecting IoT devices in home networks.}
	\vspace{-1mm}
	\label{fig:architecture}
\end{figure}

With high-level observations made in the previous section, let us assume labeled data is available from some seen contexts (say, $N$ homes). One may employ different strategies to train models and apply them to unseen contexts in practice. We illustrate in Fig.~\ref{fig:architecture} two representative strategies, highlighted by purple (global) and green (contextualized) colors. 
Suppose we have $d$ days worth of labeled data from $N$ homes, constructing our training set (shown in the left section of Fig.~\ref{fig:architecture}). Therefore, $H_{sN,d}$ denotes the unit of dataset collected from the seen home $N$ on day $d$.  

The most straightforward strategy (baseline) sounds we combine all labeled data from $N$ homes and train a classifier (shown as purple $M_g$) based on the combined dataset. We call this global modeling throughout the paper. Alternatively, one may go a step back and, instead of combing all data, trains a separate model per home (shown as green $m_1,..., m_N$). We call this method contextualized modeling. When it comes to the testing phase (detecting a set of IoT devices in an unseen home $H_u$), shown in the right section of Fig.~\ref{fig:architecture}, $M_g$ is readily applied to the daily data of $H_{u,K}$, giving prediction. For contextualized modeling, instead, an additional computation is required. It needs to ``select the best'' model from $N$ available models, before applying it to data of an unseen home.
Developing methods for the best model selection is beyond the scope of this paper and is left for our future work. Note that our primary objective is to compare the efficacy of global versus contextualized modeling, assuming the best model can be selected (either automatically or ``given''). In this paper, we leverage ground truth labels (which will not be feasible in real practice) available for all contexts (seen and unseen) during the training and testing phases of our experiments. In other words, we choose the model that yields the highest accuracy (given ground truth) for unseen homes. 

These two strategies come with key differences: (1) Suppose a labeled dataset becomes available from an unseen (or even a seen) home. Global modeling requires to re-train $M_g$ on past data combined with new data. Contextualized modeling, instead, trains an isolated model specific to a newly added/updated home, which is relatively faster and less expensive computationally; and (2) Although this paper selects the best model by leveraging ground truth labels, a practical approach like what we will explain in \S\ref{sec:dynamic} may not always be able to select the best model from a set of available models (hence, affecting the inference performance). 

\textbf{Evaluating Inference Strategies:}
For our evaluation, we assume labeled data of five\footnote{Less than five seen homes result in an insufficient amount of data.} homes is available for training (seen), and data of the remaining seven homes is used for testing (unseen). To avoid creating bias, we ran our experiments ten times, and we randomly selected five training homes for each run. 
Let us call the first month (of the entire 47 days) training period. Note that we train models on data of seen homes during the training period. We train a global model ($M_g$) and five contextualized models ($m_i$) for the chosen seen homes, as shown in Fig.~\ref{fig:architecture}. 

It is important to note that contextualized modeling requires the best model assigned to an unseen home before the testing phase. Given an unseen home, the best model (one of five $m_i$'s in our evaluation) is selected and assigned based on the highest accuracy obtained by applying $m_i$'s to the labeled data of unseen homes (which is assumed to be available in this paper).  
It is possible that the best model selected for an unseen home may not necessarily perform the best during the testing period. In fact, in nine runs (out of ten), we found at least one unseen home where the ideal model for its training period differs from that of the testing period due to temporal concept drifts discussed in \S\ref{subsec:timedrift}.

\begin{table}[t!]
	\caption{Performance of global versus contextualized modeling.}
	\vspace{-1mm}
	\renewcommand{\arraystretch}{1.2}
	\begin{adjustbox}{width=0.485\textwidth}
		\begin{tabular}{|c|c|c|c|c|}
			\hline
			Run & $M_g$ & $Best(\{m_i\})$ & $Best(M_g,\{m_i\})$ & $Best(M_g,\{m_i\})^d$\\ \hline
			\ciao{1} & 0.770 &	0.748  & 0.769 & 0.782\\ \hline
			\ciao{2} & 0.827 & 0.805 & 0.834 & 0.843 \\ \hline
			\ciao{3} & 0.823 & 0.803 & 0.816 & 0.834 \\ \hline
			\ciao{4} & 0.825 &	0.791 & 0.826 & 0.840 \\ \hline
			\ciao{5} & 0.771 & 0.728 & 0.768 & 0.783\\ \hline
			\ciao{6} & 0.822 &	0.795 & 0.820 & 0.819 \\ \hline
			\ciao{7} & 0.844 &	0.782 & 0.836 & 0.837 \\ \hline
			\ciao{8} & 0.840 &	0.787 & 0.838 & 0.836 \\ \hline
			\ciao{9} & 0.841 &	0.788 & 0.840 & 0.831  \\ \hline
			\ciao{10} & 0.826 & 0.806 & 0.820 & 0.851 \\ \hline \hline
			{\textbf{Avg.}} & {\textbf{0.818}}&	{\textbf{0.783}} & {\textbf{0.816}} & {\textbf{0.826}}\\ \hline
			
		\end{tabular}
	\end{adjustbox}
	\label{tab:runs}
	\vspace{-4mm}
\end{table}

Table~\ref{tab:runs} summarizes the prediction accuracy (averaged across testing homes) for 10 runs. 
The second and third columns show the performance of global modeling (\ie $M_g$) versus that of contextualized modeling (\ie $Best(\{m_i\})$), respectively. We observe that global modeling consistently outperforms contextualized modeling ``on average'' across all runs. However, this pattern is not necessarily present at individual home levels. Let us closely look at a representative run to draw detailed insights. Considering run \ciao{2}, homes $H_1$, $H_4$, $H_5$, $H_7$, and $H_9$ were randomly chosen as seen contexts (resulting in $m_i$'s) and therefore remaining homes $H_2$, $H_3$, $H_6$, $H_8$, $H_{11}$, $H_{10}$, and $H_{12}$ were considered unseen. For unseen home $H_2$, the global model gives an accuracy of $0.822$; however, the best selected contextualized model $m_9$ gives an accuracy of $0.860$. Similarly, in the same run, for unseen home $H_8$, the accuracy of the global model is $0.774$, but $m_9$ gives better accuracy of $0.790$.

Overall, in half of the ten runs, we found cases of unseen homes whereby the contextualized model outperforms the global model. Scaling our small-size experiment to the size of the operation of an ISP with thousands of homes, one would appreciate the whole argument that either global or contextualized modeling can be sub-optimal; hence, a better strategy is required.

\begin{figure}[t!]
	\centering
	\includegraphics[width=0.985\linewidth]{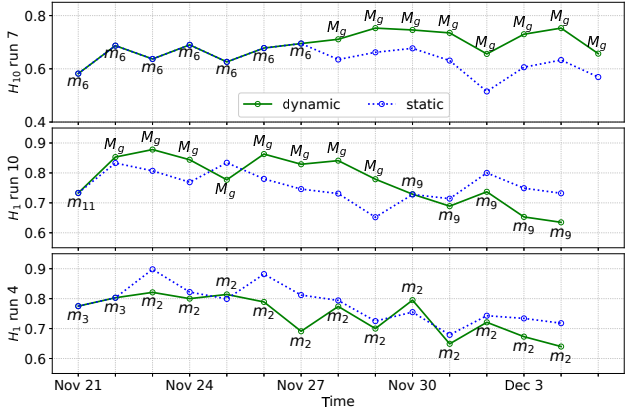}
	\vspace{-2mm}
	\caption{Accuracy of static versus dynamic models during testing period for three representative homes.}
	\label{fig:timetrace}
	\vspace{-1mm}
\end{figure}

\section{Combined and Dynamic Inference for Managing Concept Drifts}\label{sec:dynamic}

In the previous section, we saw that global modeling might sound superior at a macro level but contextualized modeling is not always worse (if not better) at a micro level. 
Therefore, we choose a strategy whereby a combined capability of global and contextualized modeling is leveraged. In other words, we bring $M_g$ as an additional context to the mix of contextualized models $\{m_i: i\in[1..N]\}$ to select the best model, $Best(Mg,\{m_i\})$, for an unseen home.  

%

Under the fourth column of Table~\ref{tab:runs} we report the accuracy of the two approaches combined.  
Unsurprisingly, the augmented contextualized modeling consistently outperforms its original form (third column of Table~\ref{tab:runs}) at a macro level (average accuracy across unseen homes).
Also, it can be seen that in some runs like \ciao{2} and \ciao{4}, the performance of the new strategy is slightly better than that of the global $M_g$. However, $M_g$ marginally wins the competition with the new strategy by the metric of aggregate accuracy. At a micro level (individual homes), we found that the two strategies yield almost the same prediction accuracy for about 60\% of testing homes across different runs. For the remaining 40\% of testing homes, the superiority of $M_g$ or the combined strategy is insignificant ($<1\%$). 
It is important to note that the best model (of the combine strategy) is selected based on observations during the training period. In other words, our selection of the best model, though it has improved by incorporating $M_g$ into the mix, is still ``static'', which makes it vulnerable to temporal concept drifts.


To overcome this challenge, we slightly refine our inference strategy and make it ``dynamic''. This means that the process of best model selection continues in the time domain (occurs, say, once a day). Such a change in our inference will certainly introduce additional computing costs (for periodic selection). Instead, the inference process becomes relatively more resilient to temporal concept drifts.   

Another practical challenge is that certain homes may not have sufficient instances (only a few IPFIX records due to the limited activity of the corresponding IoT devices) to make a meaningful selection for some days during the testing period. Therefore, we maintain a sliding window equal to the length of our testing period (a month). The data of the sliding window is considered for the dynamic selection of the best model. Note that we still have $M_g$ in our mix.   


\begin{table}[t!]
	\caption{Average number of days across 10 runs that dynamic ($d$) approach gives higher ($>$), lower ($<$), or equal ($=$) accuracy compared to static ($s$) approach.}
	\renewcommand{\arraystretch}{1}
	\begin{adjustbox}{width=0.48\textwidth}
		\begin{tabular}{|c|c|c|c|c|c|c|c|c|c|c|c|c|}
			\hline
			&  $H_1$ &  $H_2$ & $H_3$  &$H_4$  &  $H_5$  &  $H_6$  & $H_7$  & $H_8$  & $H_9$  & $H_{10}$  & $H_{11}$  &  $H_{12}$   \\ \hline
			$d>s$&  5 &  1 &  0  &  0  &  0  & 0  & 4 &  4  &  0  &  2  & 0  &  1   \\ \hline
			$d<s$ & 3 &  2 &  1  &  0  &  0  &  1  &  1  &  1  &  0  &  1  &  0 &  1  \\ \hline
			$d=s$ & 6 &  12 &  14  &  14  &  9  &  14  &  10  &  10  &  15  &  12  &  16  &  14  \\ \hline		
		\end{tabular}
	\end{adjustbox}\label{tab:days}
	\vspace{-4mm}
\end{table}

The fifth column in Table~\ref{tab:runs} shows the performance (average daily accuracy of all unseen homes) of our model selected dynamically. 
Unsurprisingly, our dynamic combined approach consistently outperforms the baseline of contextualized modeling (third column). In six (out of ten) runs, the performance of the dynamic combined approach is better than that of both $M_g$ (second column) and static combined (third column) approaches. Although $M_g$ performs slightly better in the remaining four runs, its superiority is marginal. 
Averaging across all runs (the last row in Table~\ref{tab:runs}), the dynamic approach gives an aggregate accuracy of $0.826$, which is greater than all other approaches we evaluated. {\color{black} One may argue that the dynamic approach does not offer a significant advantage compared to the global model, especially when the global model is easier to develop. This is true for our relatively small analysis across 12 home networks and 24 device types. However, at the scale of an ISP with thousands of home networks serving a larger variety of device types, the dynamic approach seems attractive, potentially outperforming the global model.}

Note that results in Table~\ref{tab:runs} are at an aggregate level. Fig.~\ref{fig:timetrace} helps us have a closer look at the performance of static versus dynamic approaches for three representative homes unseen to our models, each from a certain run, across the testing days. Blue dotted lines correspond to the prediction accuracy of the static combined approach, and green solid lines correspond to the accuracy of the dynamic combined approach. For the dynamic approach, the model selected on each day is annotated accordingly. Note that home $H_1$ was entirely inactive on 5\nth Dec, and hence no data-point for that day.

Let us start from the top plot, corresponding to home $H_{10}$ in run \ciao{7}. For the first six days (between 21\nst and 27\nth Nov), $M_6$ is the selected model by both static and dynamic approaches. However, after that, the dynamic approach selects $M_g$, which gives higher accuracy compared to $M_6$. Moving to the second plot ($H_1$ in run \ciao{10}), we observe that the dynamic approach outperforms the static one on days between 21\nst Nov and 30\nth  Nov (except for 25\nth Nov). However, its superiority fades out from 1\nst Dec onward. Lastly, for home $H_1$ in run \ciao{4}, the static approach defeats the dynamic approach almost every day, except for 25\nth Nov and 30\nth. The key takeaway is that a definite winner cannot be easily concluded from these observations.

Inspired by observations in Fig.~\ref{fig:timetrace}, we analyze the performance using a different lens. 
In each run, we count the number of days that the dynamic approach yields lower, higher, and equal accuracy compared to the static approach. 
We then compute the average count across ten runs. Table~\ref{tab:days} shows the results of this evaluation. 
While most days, both dynamic and static techniques select the same model ($D=S$), the dynamic approach outperforms when the two differ. It can be seen that in 17 home days, the dynamic approach wins ($D>S$) versus 11 home days with the static approach winning ($D<S$), meaning more than $50$\% superiority for the dynamic approach.
We also quantified the accuracy delta between these two techniques. When the dynamic approach wins, it outperforms by $0.004$, on average. This metric is about $0.002$ for the static approach, meaning half of the dynamic approach.
With the dynamic approach, we found that for $20$\% of home days, one of $\{m_i\}$ (instead of $M_g$) is selected for inference. This means our dynamic combined method improves the baseline inference (by $M_g$) in a fifth of our experimented scenarios.

\textbf{Model Selection in Absence of Labeled Data:}
As stated earlier, 
this paper assumed that the best model is somewhat given (by leveraging ground truth labels of an unseen dataset) for the contextualized modeling approach. However, in practice, a method (even an approximation) is required to guide the best model selection process. Methods like those discussed in \cite{Reis2018b} employ the classification scores of inference models to construct the data distribution of various contexts. The similarity of distributions (quantifiable by several metrics) would determine the best model (closely matching the distribution of unseen data). Suppose the score distribution of unseen testing home $H^U_i$ has the shortest distance from the score distribution of seen training home $H^s_j$. Their data distributions are likely similar, so model $m_j$ is the best candidate for inference from data of home $H^U_i$. We will thoroughly analyze this approach and corresponding metrics in our future work, developing a practical and scalable selection method for contextualized modeling.

\section{Related Work}
{\textbf{Classifying IoT devices}} in residential networks has been studied by several prior works \cite{LCN2021,Marchal19,Meidan2020,Miettinen2017,Thangavelu2019}. Works in \cite{Marchal19,Thangavelu2019,Miettinen2017} require that the network traffic is measured pre-NAT (\ie before leaving the home gateway), whereby home gateways require (software and/or hardware) modifications. Works in \cite{LCN2021,Meidan2020} measure traffic post-NAT and employ tools like IPFIX and NetFlow to generate network flows. Post-NAT data collection is proven to scale cost-effectively without requiring to instrument thousands of homes distributed across a large geography. Pre-NAT data collection, instead, provides richer visibility, particularly into end device identities like their private IP and/or MAC addresses.  

Similar to our work, authors of in \cite{LCN2021} detect (via supervised learning) IoT device types in home networks based IPFIX features. They only focused on data from a lab (single context).  
AuDI \cite{Marchal19} leverages patterns in periodic communications to determine (via unsupervised learning) clusters of IoT devices. Works in \cite{Meidan2020, Miettinen2017} train a one-vs-rest classifier for each device type --  \cite{Meidan2020} analyzes features of NetFlow records, while \cite{Miettinen2017} focuses on packet-based features. DEFT \cite{Thangavelu2019} employs a combination of packet-based and flow-based features to train a multi-class classifier and a clustering model. 

{\textbf{Contextualized modeling}} has been studied in \cite{Ying2005, Gama2009, Reis2018a, Reis2018b, Nasc2018} in domains like computer vision, insect species classification, and air quality detection. Works in \cite{Ying2005, Gama2009} assumed that true labels for testing instances would become available in the future, guiding the decision of either selecting a previously trained model (if the performance is satisfactory) or training a new model. In \cite{Reis2018a, Reis2018b}, authors explained that obtaining true labels could be delayed or even infeasible. Therefore, they developed techniques to select the best model (from a set of trained models) using statistical tests to measure the similarity between distributions without requiring labels. They showed that matching the distribution of model scores yields acceptable results while it is computationally more attractive than computing the distance in the distribution of multi-dimensional feature space.


Authors of \cite{Reis2018a, Reis2018b,Nasc2018} studied the performance of global versus contextualized modeling in isolation when applied to domains other than network traffic analysis for classification and regression tasks. 
Work in \cite{Nasc2018} showed that contextualized neural network models outperform the global model in regression, while the result was flipped in the classification problem.
Work in \cite{Reis2018b} aimed at developing relatively complex selection methods to help the contextualized approach slightly win in competition with the global approach. 
To the best of our knowledge, this paper is the first work highlighting the efficacy of global and contextualized strategies for IoT traffic inference.  Our dynamic combination of the two strategies improves the quality of traffic inference by $20$\%.

\section{Conclusion}
This paper mainly focused on the challenge of concept drifts in modeling the behavior of IoT devices in residential networks. We collected and analyzed over 6 million IPFIX records from 12 homes, each serving 24 IoT device types. We quantified the impact of concept drifts in traffic data across time and space domains. Given concept drifts, we next quantitatively compared the performance of two broad inference strategies: global (one model trained on aggregate data from all seen homes) versus contextualized (one model per seen home) when predicting traffic data of unseen homes. Finally, we dynamically combine the capabilities of the two strategies, improving the baseline inference by $20$\%. 

\vspace{10mm}
\bibliographystyle{ACM-Reference-Format}
\bibliography{ConceptDrift}


\begin{thebibliography}{23}


\ifx \showCODEN    \undefined \def \showCODEN     #1{\unskip}     \fi
\ifx \showDOI      \undefined \def \showDOI       #1{#1}\fi
\ifx \showISBNx    \undefined \def \showISBNx     #1{\unskip}     \fi
\ifx \showISBNxiii \undefined \def \showISBNxiii  #1{\unskip}     \fi
\ifx \showISSN     \undefined \def \showISSN      #1{\unskip}     \fi
\ifx \showLCCN     \undefined \def \showLCCN      #1{\unskip}     \fi
\ifx \shownote     \undefined \def \shownote      #1{#1}          \fi
\ifx \showarticletitle \undefined \def \showarticletitle #1{#1}   \fi
\ifx \showURL      \undefined \def \showURL       {\relax}        \fi
\providecommand\bibfield[2]{#2}
\providecommand\bibinfo[2]{#2}
\providecommand\natexlab[1]{#1}
\providecommand\showeprint[2][]{arXiv:#2}

\bibitem[Aronoff(2020)]%
        {minim2020}
\bibfield{author}{\bibinfo{person}{D. Aronoff}.}
  \bibinfo{year}{2020}\natexlab{}.
\newblock \bibinfo{title}{{Top 5 IoT Vulnerability Exploits in Smart Homes}}.
\newblock
\newblock
\urldef\tempurl%
\url{https://bit.ly/3f5ElXo}
\showURL{%
\tempurl}


\bibitem[Avira(2022)]%
        {aviraSafeThings}
\bibfield{author}{\bibinfo{person}{Avira}.} \bibinfo{year}{2022}\natexlab{}.
\newblock \bibinfo{title}{{Avira SafeThings}}.
\newblock
\newblock
\urldef\tempurl%
\url{https://oem.avira.com/en/solutions/safethings-for-router-manufacturers}
\showURL{%
\tempurl}


\bibitem[Bitdefender(2020)]%
        {bitdefender2020}
\bibfield{author}{\bibinfo{person}{Bitdefender}.}
  \bibinfo{year}{2020}\natexlab{}.
\newblock \bibinfo{title}{{Common IoT Devices Become the ISPs' Worst Enemy}}.
\newblock
\newblock
\urldef\tempurl%
\url{https://businessinsights.bitdefender.com/common-iot-devices-become-the-isps-worst-enemy}
\showURL{%
\tempurl}


\bibitem[Bitdefender(2022)]%
        {bitdefenderSolution}
\bibfield{author}{\bibinfo{person}{Bitdefender}.}
  \bibinfo{year}{2022}\natexlab{}.
\newblock \bibinfo{title}{{Bitdefender IoT Security Platform}}.
\newblock
\newblock
\urldef\tempurl%
\url{https://www.bitdefender.com/iot/}
\showURL{%
\tempurl}


\bibitem[Gama et~al\mbox{.}(2009)]%
        {Gama2009}
\bibfield{author}{\bibinfo{person}{J. Gama} {et~al\mbox{.}}}
  \bibinfo{year}{2009}\natexlab{}.
\newblock \showarticletitle{{Tracking Recurring Concepts with Meta-learners}}.
  In \bibinfo{booktitle}{\emph{Proc. Progress in Artificial Intelligence}}.
  \bibinfo{address}{Berlin, Germany}.
\newblock


\bibitem[Huang et~al\mbox{.}(2020)]%
        {IoTInspector}
\bibfield{author}{\bibinfo{person}{D. Huang} {et~al\mbox{.}}}
  \bibinfo{year}{2020}\natexlab{}.
\newblock \showarticletitle{{IoT Inspector: Crowdsourcing Labeled Network
  Traffic from Smart Home Devices at Scale}}.
\newblock \bibinfo{journal}{\emph{ACM IMWUT}} \bibinfo{volume}{4},
  \bibinfo{number}{2} (\bibinfo{date}{Jun} \bibinfo{year}{2020}),
  \bibinfo{pages}{1--12}.
\newblock


\bibitem[Kolcun et~al\mbox{.}(2021)]%
        {TMA2021revistIoTclass}
\bibfield{author}{\bibinfo{person}{R. Kolcun} {et~al\mbox{.}}}
  \bibinfo{year}{2021}\natexlab{}.
\newblock \showarticletitle{{Revisiting IoT Device Identification}}. In
  \bibinfo{booktitle}{\emph{Proc. IFIP TMA}}. \bibinfo{address}{Virtual}.
\newblock


\bibitem[Marchal et~al\mbox{.}(2019)]%
        {Marchal19}
\bibfield{author}{\bibinfo{person}{S. Marchal} {et~al\mbox{.}}}
  \bibinfo{year}{2019}\natexlab{}.
\newblock \showarticletitle{{AuDI: Toward Autonomous IoT Device-Type
  Identification Using Periodic Communication}}.
\newblock \bibinfo{journal}{\emph{IEEE JSAC}} \bibinfo{volume}{37},
  \bibinfo{number}{6} (\bibinfo{date}{Jun} \bibinfo{year}{2019}),
  \bibinfo{pages}{1402--1412}.
\newblock


\bibitem[Meidan et~al\mbox{.}(2020)]%
        {Meidan2020}
\bibfield{author}{\bibinfo{person}{Y. Meidan} {et~al\mbox{.}}}
  \bibinfo{year}{2020}\natexlab{}.
\newblock \showarticletitle{{A Novel Approach For Detecting Vulnerable IoT
  Devices Connected Behind a Home NAT}}.
\newblock \bibinfo{journal}{\emph{Computers \& Security}}  \bibinfo{volume}{97}
  (\bibinfo{date}{Oct} \bibinfo{year}{2020}), \bibinfo{pages}{1--23}.
\newblock


\bibitem[Miettinen et~al\mbox{.}(2017)]%
        {Miettinen2017}
\bibfield{author}{\bibinfo{person}{M. Miettinen} {et~al\mbox{.}}}
  \bibinfo{year}{2017}\natexlab{}.
\newblock \showarticletitle{{IoT SENTINEL: Automated Device-Type Identification
  for Security Enforcement in IoT}}. In \bibinfo{booktitle}{\emph{Proc. IEEE
  ICDCS}}. \bibinfo{address}{Atlanta, USA}.
\newblock


\bibitem[Nascimento et~al\mbox{.}(2018)]%
        {Nasc2018}
\bibfield{author}{\bibinfo{person}{N. Nascimento} {et~al\mbox{.}}}
  \bibinfo{year}{2018}\natexlab{}.
\newblock \showarticletitle{{A Context-Aware Machine Learning-Based Approach}}.
  In \bibinfo{booktitle}{\emph{Proc. CASCON}}. \bibinfo{address}{Markham,
  Canada}.
\newblock


\bibitem[PaloAlto(2020)]%
        {Palo2020}
\bibfield{author}{\bibinfo{person}{PaloAlto}.} \bibinfo{year}{2020}\natexlab{}.
\newblock \bibinfo{title}{{Unit 42 IoT Threat Report}}.
\newblock
\newblock
\urldef\tempurl%
\url{https://unit42.paloaltonetworks.com/iot-threat-report-2020/}
\showURL{%
\tempurl}


\bibitem[Pashamokhtari et~al\mbox{.}(2021)]%
        {LCN2021}
\bibfield{author}{\bibinfo{person}{A. Pashamokhtari} {et~al\mbox{.}}}
  \bibinfo{year}{2021}\natexlab{}.
\newblock \showarticletitle{{Inferring Connected IoT Devices from IPFIX Records
  in Residential ISP Networks}}. In \bibinfo{booktitle}{\emph{Proc. IEEE LCN}}.
  \bibinfo{address}{Virtual Event, Canada}.
\newblock


\bibitem[Reis et~al\mbox{.}(2018a)]%
        {Reis2018b}
\bibfield{author}{\bibinfo{person}{D.~M. Reis} {et~al\mbox{.}}}
  \bibinfo{year}{2018}\natexlab{a}.
\newblock \showarticletitle{{Classifying and Counting with Recurrent
  Contexts}}. In \bibinfo{booktitle}{\emph{Proc. ACM SIGKDD}}.
  \bibinfo{address}{London, United Kingdom}.
\newblock


\bibitem[Reis et~al\mbox{.}(2018b)]%
        {Reis2018a}
\bibfield{author}{\bibinfo{person}{D.~M. Reis} {et~al\mbox{.}}}
  \bibinfo{year}{2018}\natexlab{b}.
\newblock \showarticletitle{{Unsupervised Context Switch for Classification
  Tasks on Data Streams with Recurrent Concepts}}. In
  \bibinfo{booktitle}{\emph{{Proc. ACM SAC}}}. \bibinfo{address}{Pau, France}.
\newblock


\bibitem[Saidi et~al\mbox{.}(2020)]%
        {Saidi2020}
\bibfield{author}{\bibinfo{person}{S.~J. Saidi} {et~al\mbox{.}}}
  \bibinfo{year}{2020}\natexlab{}.
\newblock \showarticletitle{{A Haystack Full of Needles: Scalable Detection of
  IoT Devices in the Wild}}. In \bibinfo{booktitle}{\emph{Proc. ACM IMC}}.
  \bibinfo{address}{New York, USA}.
\newblock


\bibitem[{Sivanathan} et~al\mbox{.}(2019)]%
        {TMC19arunan}
\bibfield{author}{\bibinfo{person}{A. {Sivanathan}} {et~al\mbox{.}}}
  \bibinfo{year}{2019}\natexlab{}.
\newblock \showarticletitle{{Classifying IoT Devices in Smart Environments
  Using Network Traffic Characteristics}}.
\newblock \bibinfo{journal}{\emph{IEEE TMC}} \bibinfo{volume}{18},
  \bibinfo{number}{8} (\bibinfo{date}{Aug} \bibinfo{year}{2019}),
  \bibinfo{pages}{1745--1759}.
\newblock


\bibitem[Sivanathan et~al\mbox{.}(2020a)]%
        {detectChange2020}
\bibfield{author}{\bibinfo{person}{A. Sivanathan} {et~al\mbox{.}}}
  \bibinfo{year}{2020}\natexlab{a}.
\newblock \showarticletitle{{Detecting Behavioral Change of IoT Devices Using
  Clustering-Based Network Traffic Modeling}}.
\newblock \bibinfo{journal}{\emph{IEEE Internet of Things Journal}}
  \bibinfo{volume}{7}, \bibinfo{number}{8} (\bibinfo{date}{Aug}
  \bibinfo{year}{2020}), \bibinfo{pages}{7295--7309}.
\newblock


\bibitem[Sivanathan et~al\mbox{.}(2020b)]%
        {TNSM2021}
\bibfield{author}{\bibinfo{person}{A. Sivanathan} {et~al\mbox{.}}}
  \bibinfo{year}{2020}\natexlab{b}.
\newblock \showarticletitle{Managing IoT Cyber-Security Using Programmable
  Telemetry and Machine Learning}.
\newblock \bibinfo{journal}{\emph{IEEE TNSM}} \bibinfo{volume}{17},
  \bibinfo{number}{1} (\bibinfo{date}{Feb} \bibinfo{year}{2020}),
  \bibinfo{pages}{60--74}.
\newblock


\bibitem[Thangavelu et~al\mbox{.}(2019)]%
        {Thangavelu2019}
\bibfield{author}{\bibinfo{person}{V. Thangavelu} {et~al\mbox{.}}}
  \bibinfo{year}{2019}\natexlab{}.
\newblock \showarticletitle{{DEFT: A Distributed IoT Fingerprinting
  Technique}}.
\newblock \bibinfo{journal}{\emph{IEEE IoTJ}} \bibinfo{volume}{6},
  \bibinfo{number}{1} (\bibinfo{date}{Feb} \bibinfo{year}{2019}),
  \bibinfo{pages}{940--952}.
\newblock


\bibitem[Trammell et~al\mbox{.}(2008)]%
        {ipfixRFC5103}
\bibfield{author}{\bibinfo{person}{B. Trammell} {et~al\mbox{.}}}
  \bibinfo{year}{2008}\natexlab{}.
\newblock \bibinfo{title}{{Bidirectional Flow Export Using IP Flow Information
  Export (IPFIX)}}.
\newblock \bibinfo{howpublished}{RFC 5103}.
\newblock
\urldef\tempurl%
\url{https://doi.org/10.17487/RFC5103}
\showDOI{\tempurl}


\bibitem[Yang et~al\mbox{.}(2005)]%
        {Ying2005}
\bibfield{author}{\bibinfo{person}{Y. Yang} {et~al\mbox{.}}}
  \bibinfo{year}{2005}\natexlab{}.
\newblock \showarticletitle{{Combining Proactive and Reactive Predictions for
  Data Streams}}. In \bibinfo{booktitle}{\emph{Pro. ACM SIGKDD}}.
  \bibinfo{address}{Chicago, USA}.
\newblock


\bibitem[Zhao et~al\mbox{.}(2021)]%
        {zhao21}
\bibfield{author}{\bibinfo{person}{J. Zhao} {et~al\mbox{.}}}
  \bibinfo{year}{2021}\natexlab{}.
\newblock \showarticletitle{{Network Traffic Classification for Data Fusion: A
  Survey}}.
\newblock \bibinfo{journal}{\emph{Information Fusion}}  \bibinfo{volume}{72}
  (\bibinfo{year}{2021}), \bibinfo{pages}{22--47}.
\newblock
\showISSN{1566-2535}


\end{thebibliography}

\end{document}